\documentclass[sigconf]{acmart}

\copyrightyear{2020}
\acmYear{2020}
\setcopyright{acmcopyright}\acmConference[MM '20]{Proceedings of the 28th ACM
International Conference on Multimedia}{October 12--16, 2020}{Seattle, WA, USA}
\acmBooktitle{Proceedings of the 28th ACM International Conference on Multimedia
(MM '20), October 12--16, 2020, Seattle, WA, USA}
\acmPrice{15.00}
\acmDOI{10.1145/3394171.3413630}
\acmISBN{978-1-4503-7988-5/20/10}

\usepackage{indentfirst}
\begin{document}
\fancyhead{}

\acmSubmissionID{1329}



\title{SimSwap: An Efficient Framework For High Fidelity\\Face Swapping}

\author{Renwang Chen$^{1*}$, Xuanhong Chen$^{1*}$, Bingbing Ni$^{1,\boxtimes}$ and Yanhao Ge$^{2}$}
\thanks{$^\ast$Equal contribution.}
\thanks{$^\boxtimes$Corresponding author: Bingbing Ni.}
\affiliation{%
	\institution{$^1$Shanghai Jiao Tong University, Shanghai, China}
	\institution{$^2$Tencent, China}
}
\email{{applebananac,chen19910528,nibingbing}@sjtu.edu.cn, halege@tencent.com}

\renewcommand{\thefootnote}{\fnsymbol{footnote}}
\renewcommand{\shortauthors}{Chen and Chen, et al.}

\begin{abstract}
We propose an efficient framework, called \emph{Simple Swap (SimSwap)}, aiming for generalized and high fidelity face swapping. In contrast to previous approaches that either lack the ability to generalize to arbitrary identity or fail to preserve attributes like facial expression and gaze direction, our framework is capable of transferring the identity of an arbitrary source face into an arbitrary target face while preserving the attributes of the target face. We overcome the above defects in the following two ways. First, we present the \emph{ID Injection Module (IIM)} which transfers the identity information of the source face into the target face at feature level. By using this module, we extend the architecture of an identity-specific face swapping algorithm to a framework for arbitrary face swapping. Second, we propose the \emph{Weak Feature Matching Loss} which efficiently helps our framework to preserve the facial attributes in an implicit way. Extensive experiments on wild faces demonstrate that our \emph{SimSwap} is able to achieve competitive identity performance while preserving attributes better than previous state-of-the-art methods.
The code is already available on github: \url{https://github.com/neuralchen/SimSwap}.
\end{abstract}


\begin{teaserfigure}
  \includegraphics[width=\textwidth]{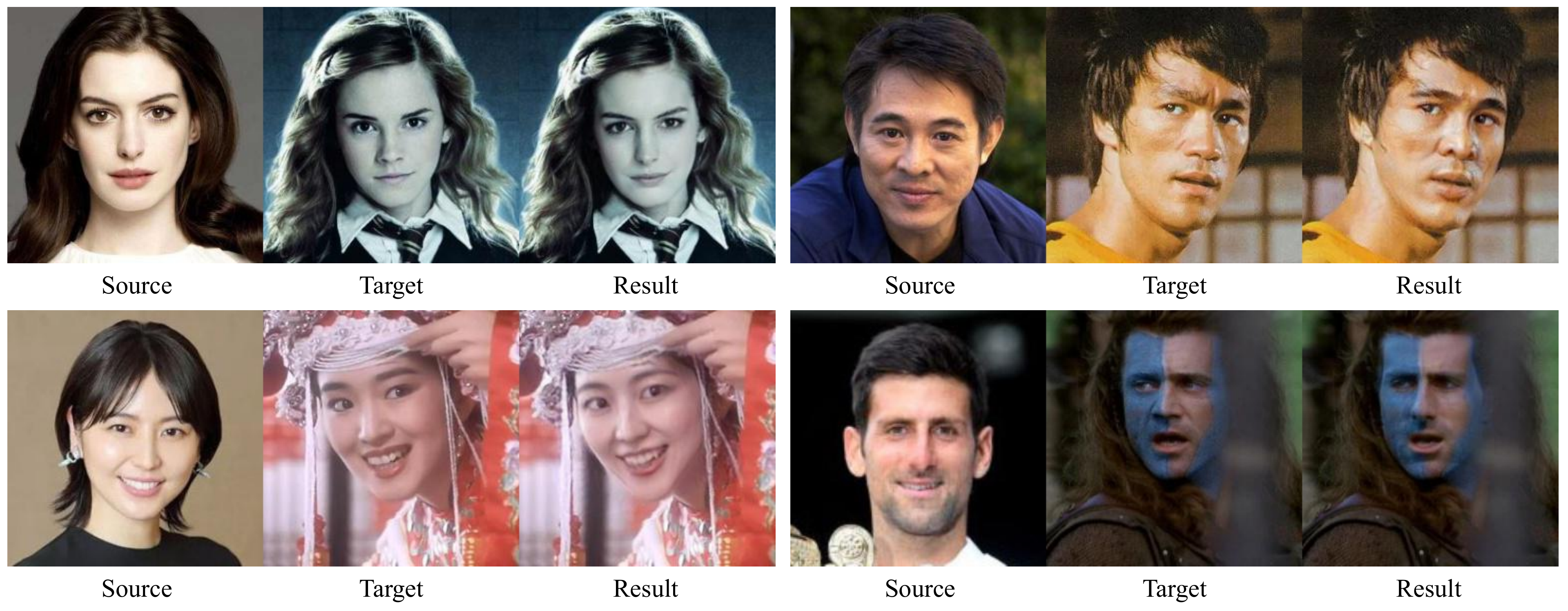}
  \caption{Face swapping results generated by SimSwap. We replace the face in the target image with the face in the source image. More results can be found in supplementary material.}
  \Description{The motivation figure of our framework.}
  \label{fig:teaser}
\end{teaserfigure}

\maketitle

\section{Introduction}
Face swapping is a promising technology that transfers the identity of a source face into a target face while keeping the attributes (\emph{e.g.} expression, posture, lighting etc.) of the target face unchanged. It has been widely used in film industry to produce nonexistent twins. The industrial face swapping method utilizes advanced equipment to reconstruct the actor's face model and rebuild the scene's lighting condition, which is beyond the reach of most people. Recently, face swapping without high-end equipment~\cite{DBLP:Identity,DBLP:FaceShifter,DBLP:FSGAN,DeepFakes} has attracted the researcher's attention.

The main difficulties in face swapping can be concluded as follows:
1). A face swapping framework with a strong generalization ability should be adapted to arbitrary faces;
2). The identity of the result face should be close to the identity of the source face;
3). The attributes(\emph{e.g.} expression, posture, lighting etc.) of the result face should be consistent with the attributes of the target face.

There are mainly two types of face swapping methods, including source-oriented methods that work on the source face at image level and target-oriented methods that work on the target face at feature level.
Source-oriented methods~\cite{DBLP:Blanz,DBLP:Dimitri,DBLP:Nirkin,DBLP:FSGAN} transfer attributes(like expression and posture) from the target face to the source face and then blend the source face into the target image. These methods are sensitive to the posture and lighting of the source image and are not able to reproduce the target's expression accurately.
Target-oriented approaches~\cite{DeepFakes,DBLP:Korshunova,DBLP:Identity,DBLP:FaceShifter} directly modify the features of the target image and can be well adapted to the variation of the source face. 
The open-source algorithm~\cite{DeepFakes} is able to generate face swapping results between two specific identities, but lacks the ability for generalization. The GAN-based work~\cite{DBLP:Identity} combines the source's identity and the target's attributes at the feature level and extends the application to arbitrary identity. A recent work~\cite{DBLP:FaceShifter} utilizes a two-stage framework and achieves high fidelity results. However, these methods focus too much on identity modification. They apply weak constrain on attribute preservation and often encounter mismatch in expression or posture.

To overcome the defects in generalization and attribute preservation, we propose an efficient face swapping framework, called \emph{SimSwap}. We analyze the architecture of an identity-specific face swapping algorithm~\cite{DeepFakes} and find out the lack of generalization is caused by the integration of identity information into the Decoder so the Decoder can be only applied to one specific identity. To avoid such integration, we present the \textit{ID Injection Module}. Our module conducts modifications on the features of the target image by embedding the identity information of the source face, so the relevance between identity information and the weights of Decoder can be removed and our architecture can be applied to arbitrary identities. Furthermore, identity and attribute information are highly coupled at feature level. A direct modification on the whole features will lead to a decrease in attribute performance and we need to use training losses to alleviate the effect. While explicitly constraining each attribute of the result image to match that of the target image is too complicated, we propose the \textit{Weak Feature Matching Loss}. Our Weak Feature Matching Loss aligns the generated result with the input target at high semantic level and implicitly helps our architecture to preserve the target's attributes. By using this term, our SimSwap is capable of achieving competitive identity performance while possessing a better attribute preservation skill than previous state-of-the-art methods. Extensive experiments demonstrate the generalization and effectiveness of our algorithm.

\section{Related Work}

Face swapping has been studied for a long period. The methods can be mainly divided into two types including source-oriented methods that work on the source face at image level and target-oriented methods that work on the target face at feature level.

\noindent\textbf{Source-oriented Methods.} Source-oriented methods transfer attributes from the target face to the source face and then blend the source face into the target image. Early method~\cite{DBLP:Blanz} used 3D models to transfer postures and lighting but required manual intervention. An automatic method~\cite{DBLP:Dimitri} was proposed but could only swap faces with identities in a specific face library. Nirkin et al.~\cite{DBLP:Nirkin} utilized a 3D face dataset~\cite{DBLP:BFM} to transfer the expression and posture and then used Poisson Blending~\cite{DBLP:Poisson} to merge the source face into the target image. However, owing to the limited expressiveness of the 3D face dataset, methods~\cite{DBLP:Blanz,DBLP:Nirkin} replying on 3D models often failed to reproduce the expressions accurately. Recently, FSGAN~\cite{DBLP:FSGAN} proposed a two-stage architecture which first conduct the expression and posture transfer with a face reenactment network and then used another face inpainting network to blend the source face into the target image. A common problem for source-oriented methods is that they are sensitive to the input source image. Exaggerated expression or large posture of the source face will strongly affect the performance of the face swapping result.
 
\noindent\textbf{Target-oriented Methods.} Target-oriented methods use neural network to extract the features of the target image, then conduct modifications on the features and restore the features to the output face swapping image. Korshunova et al.~\cite{DBLP:Korshunova} trained a generator and was able to swap faces with one specific identity. The famous algorithm DeepFakes~\cite{DeepFakes} utilized an Encoder-Decoder architecture. Once trained, it was able to swap faces between two specific identities but lacked the ability for generalization. Methods like ~\cite{DBLP:RSGAN,DBLP:FSNet} combined the latent representations from the source face area and target non-face area to produce the result but failed to keep target's expression. IPGAN~\cite{DBLP:Identity} extracted the identity vector from the source image and the attribute vector from the target image before sending them to the Decoder. The generated outputs were good at transferring the identity of source face but often failed to preserve the expression or posture of the target face. The recently proposed method FaceShifter~\cite{DBLP:FaceShifter} was able to produce high fidelity face swapping results. FaceShifter leveraged a sophisticated two-stage framework and achieved the-state-of-art identity performance. However, like previous methods, FaceShifter imposed too weak constrain on the attributes so their results often suffered from expression mismatch.

\begin{figure*}[!ht] 
\centering
\includegraphics[width=1\textwidth]{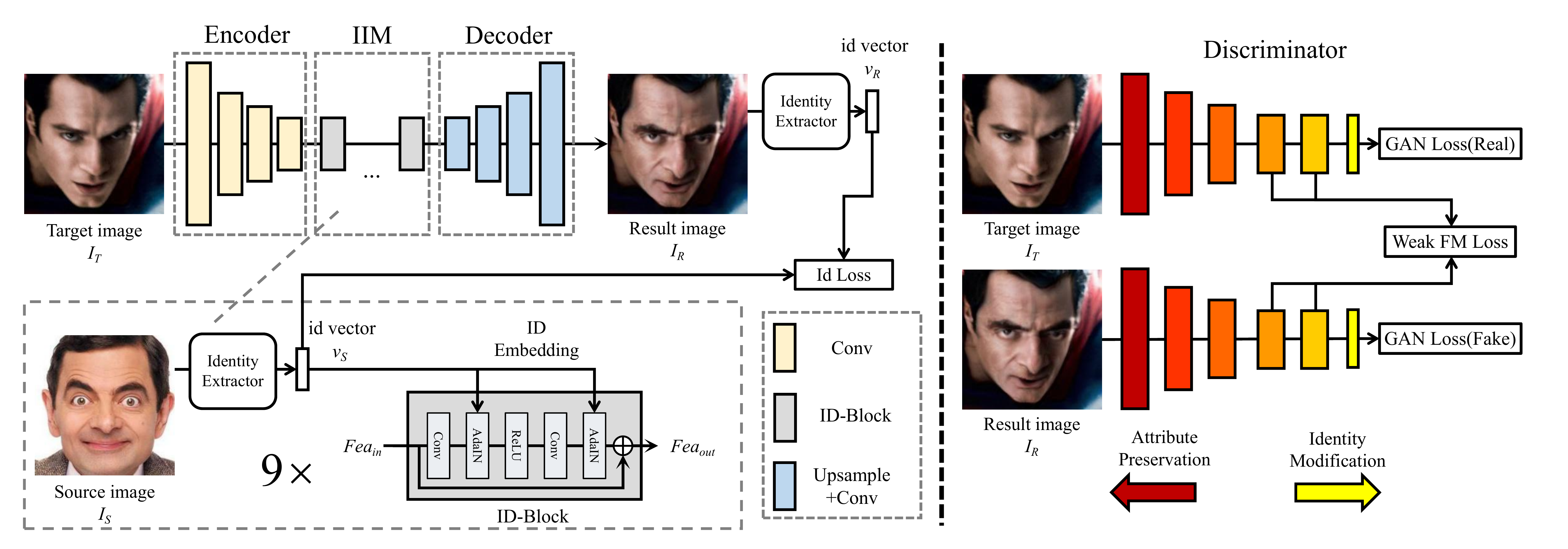}
\caption{The framework of SimSwap. Our generator consists of three parts, including the Encoder part, the ID Injection Module(IIM) and the Decoder part. The Encoder extracts features $Fea_{T}$ from the target image $I_{T}$. The ID Injection Module transfers the identity information from $I_{S}$ into $Fea_{T}$. The Decoder restores the modified features to the result image. We use Identity Loss to encourage our network to generate results with similar identity as the source face. We apply the Weak Feature Matching Loss to make sure that our network can preserve the attributes of the target face while not harming the identity modification performance too much.}
\label{fig:framework}
\end{figure*}

\section{Method}

Given a source image and a target image, we present a framework that transfers the identity of the source face into the target face while keeping the attributes of target face unchanged. Our framework extends from an identity-specific face swapping architecture~\cite{DeepFakes} and can be adapted to arbitrary identities. We first discuss the limitation of the original architecture(Sec 3.1). We show how to extend it to a framework for arbitrary identity(Sec 3.2). Then we present the Weak Feature Matching Loss which helps to preserve target's attributes(Sec 3.3). Finally, we give out our loss function(Sec 3.4).

\subsection{Limitation of the DeepFakes}

The architecture of DeepFakes contains two parts, a common Encoder $Enc$ and two identity-specific Decoders $Dec_{S}, Enc_{T}$. In the training stage, the $Enc$-$Dec_{S}$ architecture takes in the warped source images and restore them to the original unwarped source images. The same procedure will be conducted with the target images using the $Enc$-$Dec_{T}$ architecture. In the test stage, a target image will be sent to the $Enc$-$Dec_{S}$ architecture. The architecture will mistake it for a warped source image and produce an image with the source's identity and the target's attributes. 

During this process, the Encoder \textit{Enc} extracts the target's features which contain both identity and attribute information of the target face. Since the Decoder $Dec_{S}$ manages to convert the target's features to an image with source's identity, the identity information of the source face must have been integrated into the weights of $Dec_{S}$. So the Decoder in DeepFakes can be only applied to one specific identity. 

\subsection{Generalization to Arbitrary Identity}
To overcome such limitation, we are seeking a way to separate the identity information from the Decoder so that the whole architecture can be generalized to arbitrary identity. We improve the architecture by adding an additional ID Injection Module between the Encoder and the Decoder. Our framework is shown in Figure \ref{fig:framework}. Given a target image $I_{T}$, we pass it through our Encoder to extract its features $Fea_{T}$. Since our task is to swap the target face with the source face, we have to replace the identity information in $Fea_{T}$ with the identity information of source face while keeping the attribute information in $Fea_{T}$ unchanged. However, the identity and attribute information in $Fea_{T}$ are highly coupled and difficult to tell apart. So we directly conduct modifications on the whole $Fea_{T}$ and we are using the training loss to encourage the network to learn implicitly which part of $Fea_{T}$ should be changed and which part should be preserved.

Our ID Injection Module works on changing the identity information in $Fea_{T}$ towards the identity information of the source face. The module is composed of two parts, the identity extraction part and the embedding part. In the identity extraction part, we deal with the input source image $I_{S}$ which contains both identity and attribute information of source face. Since we only need the former, we use a face recognition network~\cite{DBLP:ArcFace} to extract the identity vector $v_{S}$ from $I_{S}$. In the embedding part, we are using the ID-Blocks to inject the identity information into the features. Our ID-Block is a modified version of the Residual Block~\cite{DBLP:ResNet} and we are using the Adaptive Instance Normalization(AdaIN)~\cite{DBLP:AdaIN} to replace the original Batch Normalization~\cite{DBLP:BN}. The formulation of AdaIN in our task can be written as:
\begin{equation}
  AdaIN(Fea,v_{S})=\sigma_{S}\frac{Fea-\mu(Fea)}{\sigma(Fea)}+\mu_{S}
\end{equation}
Here, $\mu(Fea)$ and $\sigma(Fea)$ is the channel-wise mean and standard deviation of the input feature $Fea$. $\sigma_{S}$ and $\mu_{S}$ are two variables generated from $v_{S}$ using full connected layers. To guarantee enough identity embedding, we are using a total of 9 ID-Blocks.

After the injection of identity information, we pass the modified features through the Decoder to generate the final result $I_{R}$. Since source images from different identities are involved in the training, the weights of the Decoder should be unrelated to any specific identity. Our Decoder will just focus on restoring the image from the features and leave the identity modification mission to the ID Injection Module, so we can apply our architecture to arbitrary identities.

During the training process, we extract the identity vector $v_{R}$ from the generated result $I_{R}$ and we use the Identity Loss to minimize the distance between $v_{R}$ and $v_{S}$. However, the minimization of the Identity Loss can make the network overfitted and only generate front face images with the source's identity while losing all the target's attributes. To avoid such phenomena, we utilize the idea of adversarial training~\cite{DBLP:GAN,DBLP:SPADE,DBLP:FUNIT,DBLP:styleGAN} and use the Discriminator to distinguish results with apparent error. The Adversarial Loss also plays an import role in improving the quality of the generated result. We use the patchGAN~\cite{DBLP:pix2pix} version of the Discriminator.

\subsection{Preserving the Attributes of the Target}
In the face swapping task, the modification should be only conducted in the identity part and the attributes (\emph{e.g.} expression, posture, lighting etc.) of the target face should remain unchanged. However, since we are directly conducting modifications on the whole $Fea_{T}$ which contains both identity and attribute information of target face, the attribute information is likely to be affected by the identity embedding. To prevent the attribute mismatch, we are using the training loss to constrain them. However, if we choose to constrain all the attributes explicitly, we will have to train one network for each attribute. The whole process should be impractical since there are too many attributes should be considered. So we propose to use the Weak Feature Matching Loss to do the constraining in an implicit way.

The idea of Feature Matching originated in pix2pixHD~\cite{DBLP:pix2pixHD} which used the Discriminator to extract multiple layers of features from the Ground Truth image and the generated output. The original Feature Matching Loss is written as:
\begin{equation}
  L_{oFM}(D) =\sum_{i=1}^{M} \frac{1}{N_{i}}\Vert D^{(i)}(I_{R})-D^{(i)}(I_{GT})\Vert_{1}
\end{equation}
Here $D^{(i)}$ denotes the $i$-th layer feature extractor of Discriminator $D$ and $N_{i}$ denotes the number of elements in the $i$-th layer. $M$ is the total number of layers. $I_{R}$ is the generated output and $I_{GT}$ is its corresponding Ground Truth image.

In our architecture, since there's no Ground Truth in face swapping task, we are using the input target image $I_{T}$ to replace its position. We remove the first few layers and only use the last few layers to calculate our Weak Feature Matching Loss, which can be written as:
\begin{equation}
  L_{wFM}(D) =\sum_{i=m}^{M} \frac{1}{N_{i}}\Vert D^{(i)}(I_{R})-D^{(i)}(I_{T})\Vert_{1}
\end{equation}
Here $m$ is the layer where we start to calculate the Weak Feature Matching Loss.

Although the original Feature Matching Loss and Weak Feature Matching Loss share similar formulations, their objectives are totally different. The original Feature Matching Loss is proposed to stabilize the training and the generator is required to produce natural statistics at multiple levels. The features of shallow layers will play the key role since they mainly contain texture information and are able to constrain the results at pixel level. However, in our face swapping task, introducing too much texture information from the input target image will make the result similar to target face and cause difficulty in identity modification, so we remove the first few layers in the original Feature Matching term. Our objective is to constrain the attribute performance. Since the attribute is high semantic information which mainly lies in deep features, we are requiring the result image to align with the input target at deep level, and our Weak Feature Matching Loss is only using the last few layers of the Discriminator to calculate the Feature Matching term. By using such a loss function, even if we are not explicitly constraining the network on any specific attribute, it will implicitly learn how to preserve the attributes of the input target face.


\subsection{Overall Loss Function}

Our Loss function has 5 components, including Identity Loss, Reconstruction Loss, Adversarial Loss, Gradient Penalty and Weak Feature Matching Loss.

\noindent\textbf{Identity Loss} Identity Loss is used to constrain the distance between $v_{R}$ and $v_{S}$. We are using the cosine similarity to calculate the distance, which can be written as:
\begin{equation}
  L_{Id} = 1-\frac{v_{R}\cdot v_{S}}{\Vert v_{R}\Vert_2\Vert v_{S}\Vert_2}
\end{equation}

\noindent\textbf{Reconstruction Loss} If the source face and the target face are from the same identity, the generated result should look the same as target face. We are using the Reconstruction Loss as a regularization term, which can be written as:
\begin{equation}
  L_{Recon} = \Vert I_{R}-I_{T}\Vert_{1}
\end{equation}
We set this term to 0 if the source and target faces are from different identities.

\noindent\textbf{Adversarial Loss and Gradient Penalty}  We are using the Hinge version~\cite{DBLP:SPADE,DBLP:BigGAN,DBLP:FUNIT} of the Adversarial Loss. We use multi-scale Discriminators~\cite{DBLP:pix2pixHD} for better performance under large postures. We also utilize the Gradient Penalty term~\cite{DBLP:WGAN,DBLP:WGAN-GP} to prevent the Discriminators from gradient explosion.

\noindent\textbf{Weak Feature Matching Loss} Since we are using mutli-scale Discriminator, the Weak Feature Matching Loss should be calculated using all Discriminators, which can be written as:
\begin{equation}
  L_{wFM\_sum}= \sum_{i=1}^2 L_{wFM}(D_{i})
\end{equation}

The overall Loss can be written as:
\begin{equation}
  \lambda_{Id}L_{Id}+\lambda_{Recon}L_{Recon}+L_{Adv}+\lambda_{GP}L_{GP}+\lambda_{wFM}L_{wFM\_sum}
\end{equation}
Here $\lambda_{Id}=10,\lambda_{Recon}=10,\lambda_{GP}=10^{-5},\lambda_{wFM}=10$.

\begin{figure*}[!ht] 
\centering
\includegraphics[width=1\textwidth]{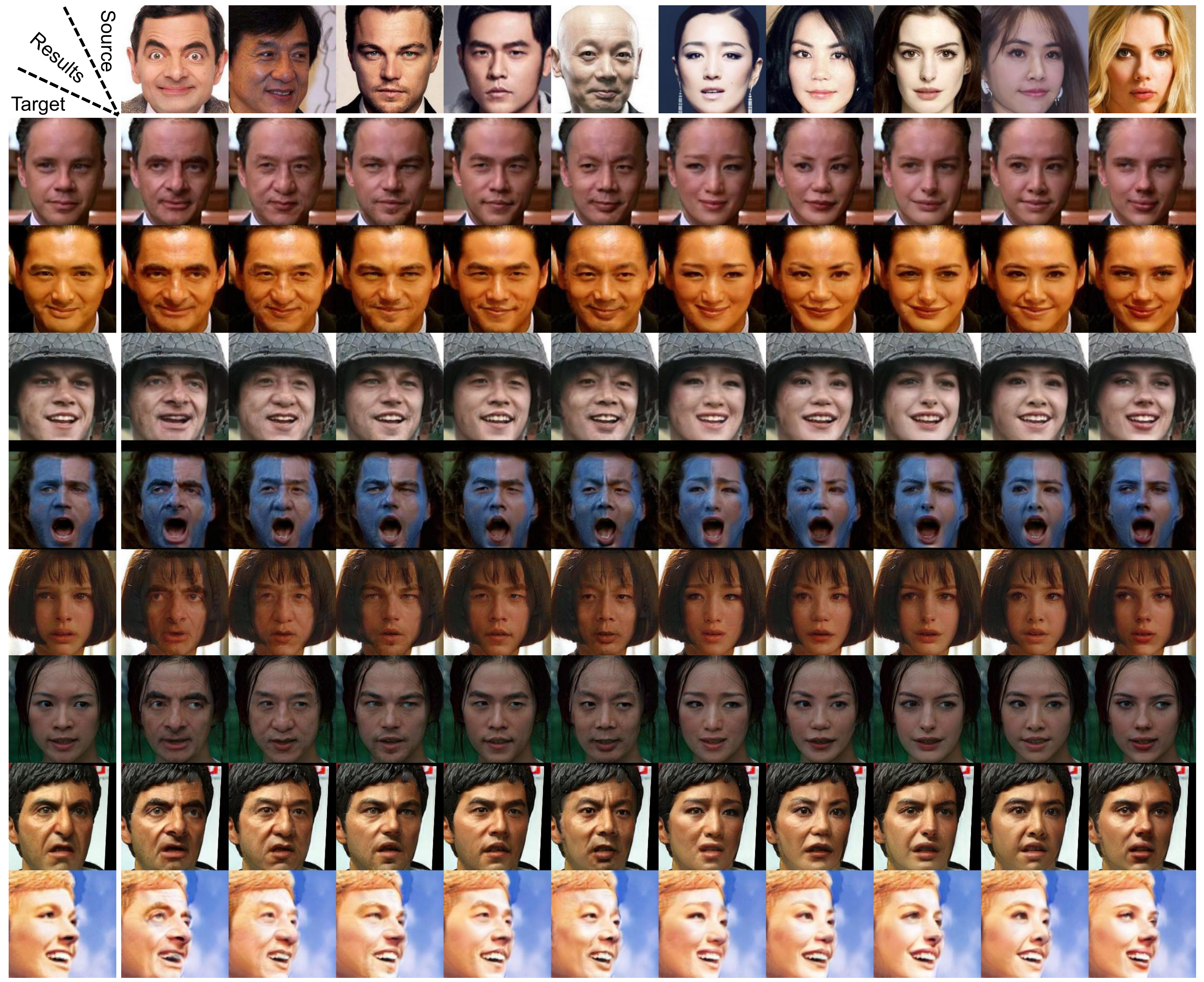}
\caption{Face matrix generated by SimSwap. The target images are picked from the movie scenes and the source images are downloaded from the Internet. All the target and source images are excluded from the training set. Source images with a front posture or neutral expression are not necessary since we are only using their identity vectors. The results show that SimSwap is able to change the identity into the source face while preserving the attributes of the target face.}
\label{fig:Matrix}
\end{figure*}

\begin{figure}
\centering
\includegraphics[width=8.5cm]{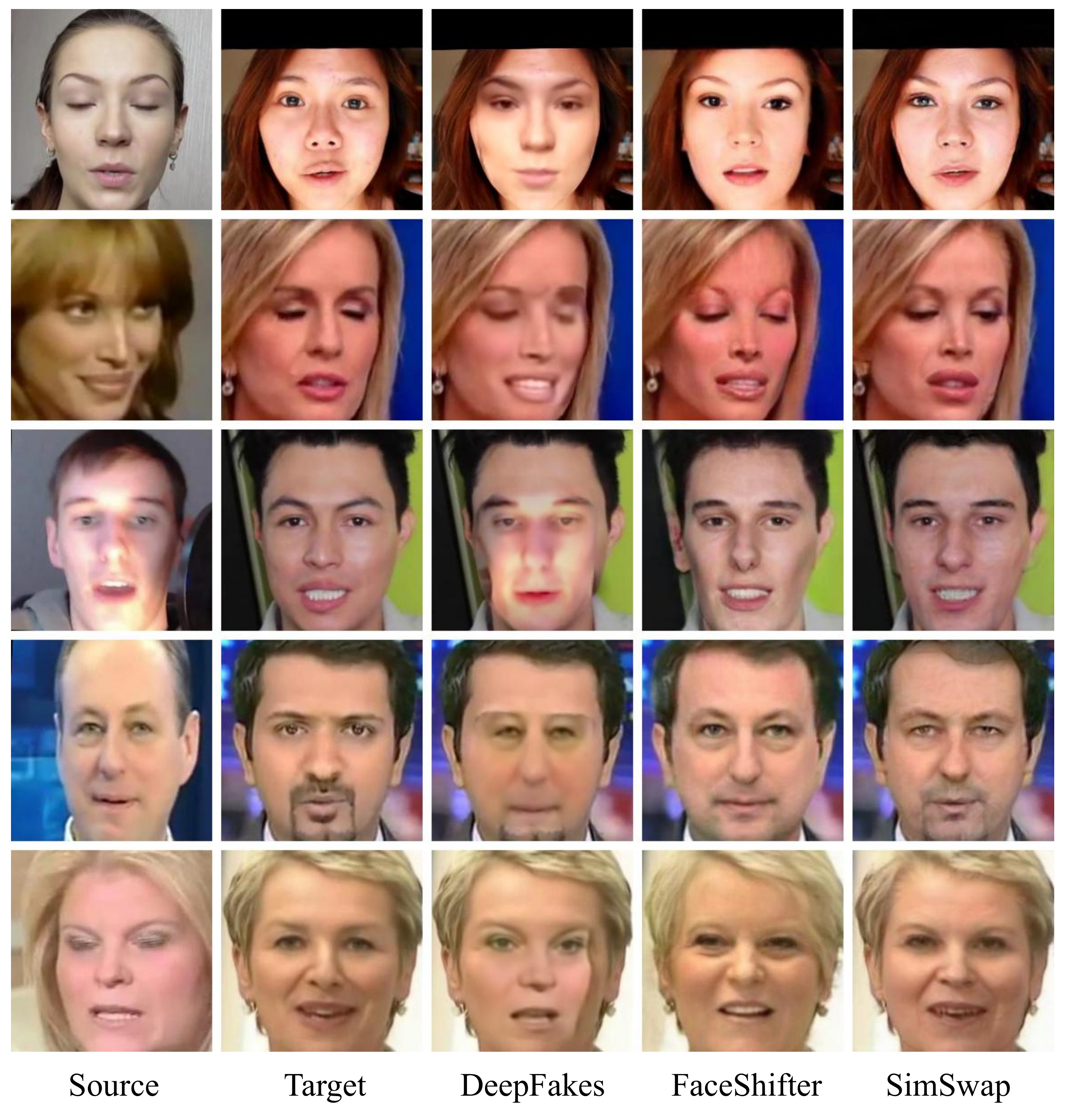}
\caption{Comparison with DeepFakes~\cite{DeepFakes} and FaceShifter~\cite{DBLP:FaceShifter} on FaceForensics++~\cite{DBLP:faceforensics}. The results of SimSwap achieve competitive identity performance while preserving better attributes (like expression, posture, lighting).}
\label{fig:DeepFakes}
\end{figure}

\section{Experiments}
\textbf{Implementation Detail} Since we are working on face swapping for arbitrary identities, we choose a large face dataset VGGFace2~\cite{DBLP:vggface} as our training set. To improve the quality of our training set, we remove images with size smaller than $250\times 250$. We align and crop the images to a standard position with size $224 \times 224$. As for the face recognition model in the ID Injection Module, we use a pretrained Arcface~\cite{DBLP:ArcFace} model on ~\cite{ArcFace_github}. We train our network using the Adam optimizer~\cite{DBLP:Adam} with $\beta_{1}=0$ and $\beta_{2}=0.999$. We train one batch for image pairs with the same identity and another batch for image pairs with different identities alternately. We train our networks for more than 500 epochs.

\subsection{Qualitative Face Swapping Results}

We are presenting a face matrix to show our face swapping results. We pick 8 face images from the movie scenes as target images. We download 10 face images from the Internet as source images. The source images are not required to have a front posture or neutral expression since we are only using their identity vectors. All these images are excluded from our training set. We conduct face swapping for all source and target pairs.


As shown in Figure \ref{fig:Matrix}, SimSwap is capable of transferring the identity of the source face into the target face while preserving the attributes (like expression, gaze direction, posture and lighting condition) of the target face. Our method can well handle various identities. Even given difficult target conditions like exaggerated expression(row 4), face stripe(row 4), large face rotations(row 7), SimSwap can still produce high fidelity face swapping results. 

\subsection{Comparison with Other Methods}

Many face swapping methods have been proposed in recent years. We choose three of them, including the source-oriented method \textbf{FSGAN}~\cite{DBLP:FSGAN} and target-oriented methods \textbf{DeepFakes}~\cite{DeepFakes}, \textbf{FaceShfiter}~\cite{DBLP:FaceShifter}. We show the comparison between these methods and our SimSwap.

\noindent\textbf{Comparison on FaceForensics++} The FaceForensics++~\cite{DBLP:faceforensics} dataset contains 1,000 face videos downloaded from Internet and 1,000 face swapping videos generated by DeepFakes. We compare SimSwap with DeepFakes and FaceShifter on FaceForensics++. Since FaceShifter doesn't release their code, we directly crop the images from their paper for fair comparison. As shown in Figure \ref{fig:DeepFakes}, the results of DeepFakes suffer from severe lighting and posture mismatch. FaceShifter manages to produce decent face swapping results but the expression and gaze direction of the result faces do not fully respect those of the target faces. Our SimSwap generates plausible face swapping results while achieving better performance in attribute preservation.

\begin{figure}
\centering
\includegraphics[width=8.5cm]{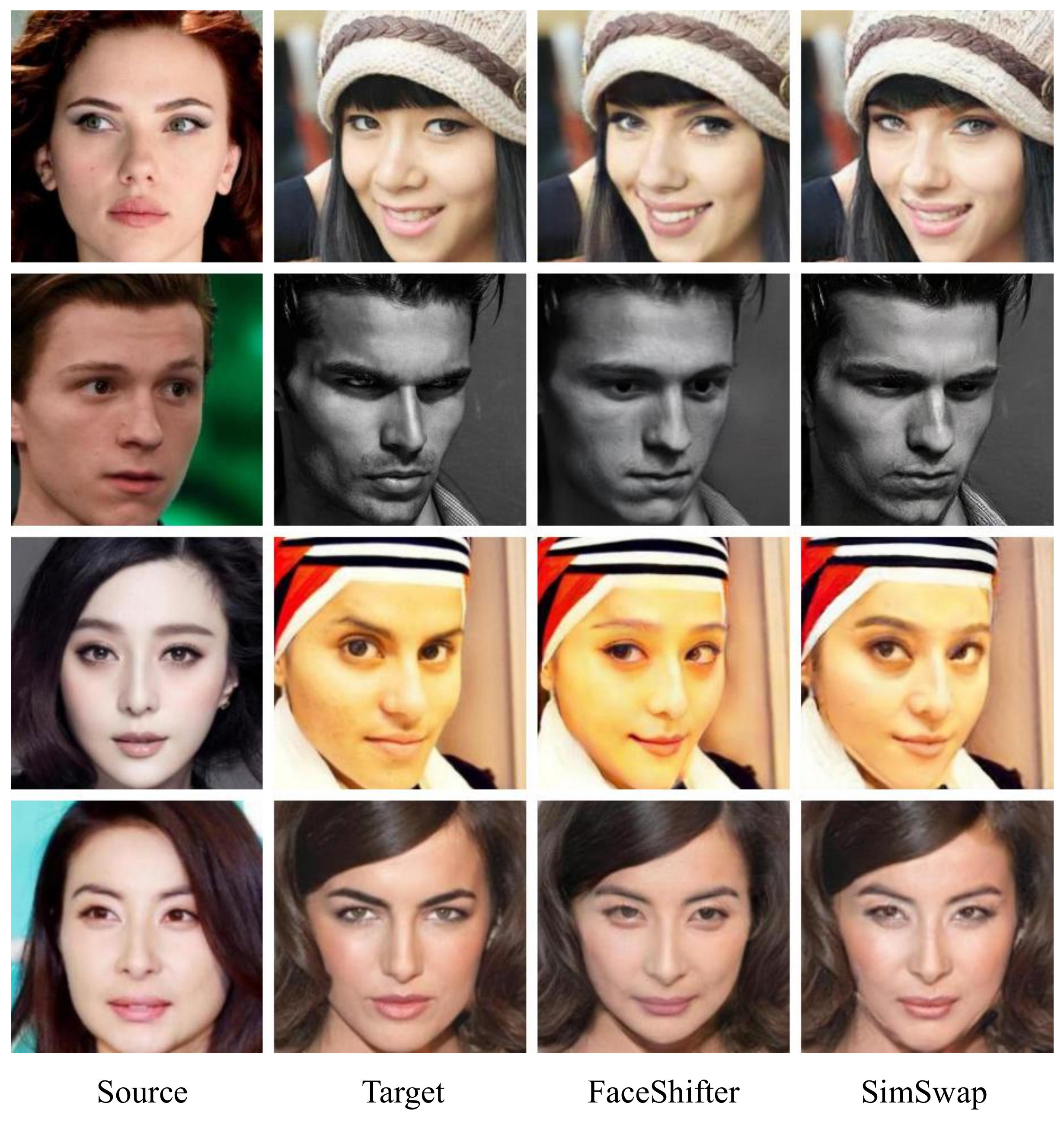}
\caption{More comparison results with FaceShifter~\cite{DBLP:FaceShifter}. We have better attribute performance in expression and lighting condition.}
\label{fig:Faceshifter}
\end{figure}

\noindent\textbf{Additional Comparison with FaceShifter} We further compare more results with FaceShifter in Figure \ref{fig:Faceshifter}. As we can see, FaceShifter exhibits a strong identity modification ability and it is able to change the face shape of the result towards that of the source face. However, it focus too much on the identity part and often fail in keeping attributes like expression and gaze direction. In Figure \ref{fig:Faceshifter} row 2, the target face is narrowing his eyes. SimSwap can generate result that reproduces such subtle expression while FaceShifter fails. Furthermore, although FaceShifter is using a second network to combine its face swapping results with the background, we still manage to produce a slightly better lighting condition(row 3\&4) than FaceShifter.

\begin{figure}
\centering
\includegraphics[width=8.5cm]{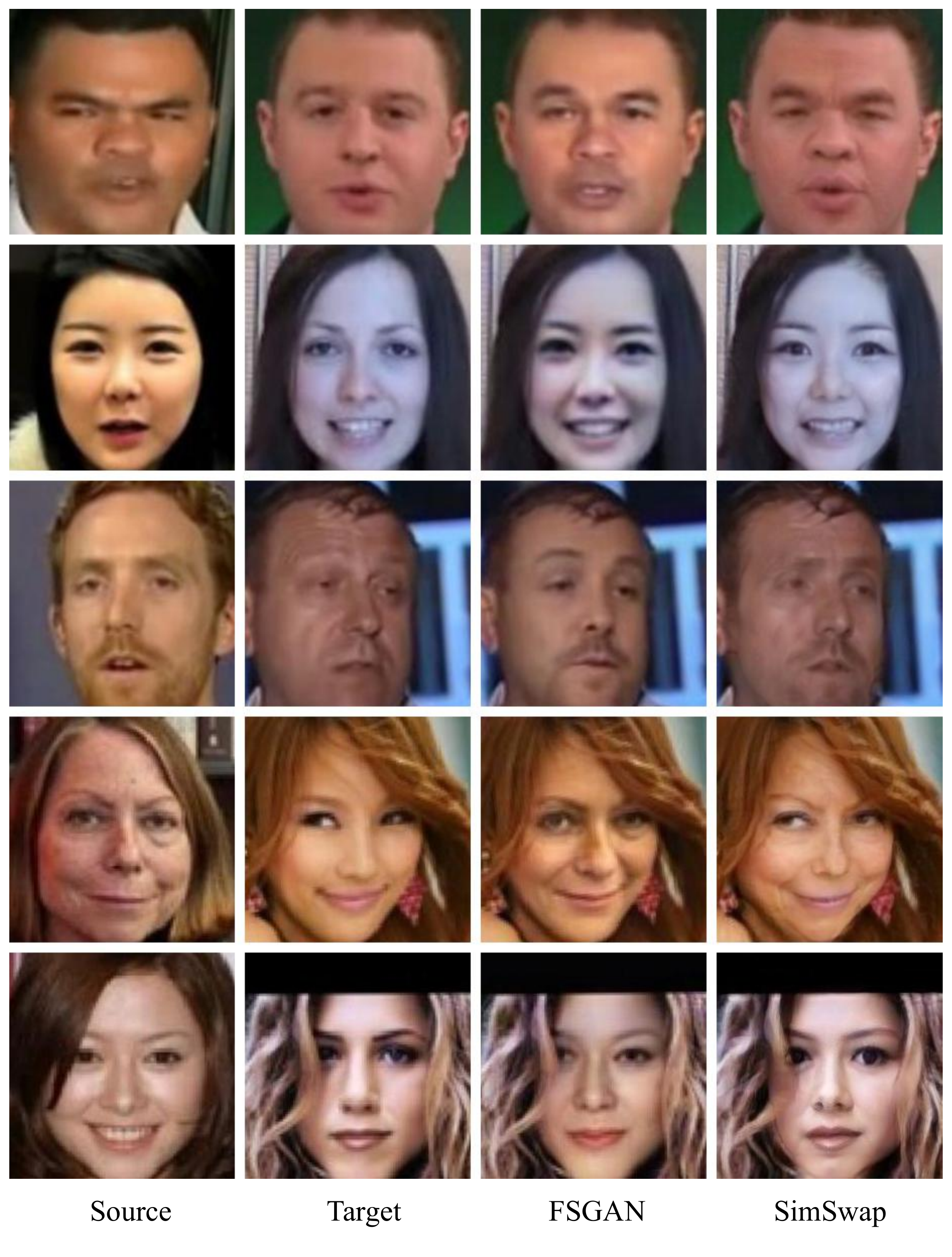}
\caption{Comparison with FSGAN~\cite{DBLP:FSGAN}. SimSwap can preserve attributes(like expression, gaze direction, lighting) better than FSGAN. Our results are less likely to be affected by the attributes of the input source images.}
\label{fig:FSGAN}
\end{figure}

\noindent\textbf{Comparison with FSGAN} We compare with FSGAN in Figure \ref{fig:FSGAN}. The results of FSGAN fail to reproduce the expression(row 1), gaze direction(row 1\&4) of the target face and there are apparent differences in lighting conditions between their results and target images. Our SimSwap achieves better performance in attribute preservation. Besides, FSGAN is very sensitive to input source image. As shown in row 2, the target face has a clear eye area but FSGAN brings in the shadows from the source face. Similar problem also occurs in row 4 around the nose area. SimSwap is much more robust to the input source image and produce more convincing results.

\subsection{Analysis of SimSwap}

In this section, we will first give an analysis of our identity modification ability. Then we will conduct several ablation tests to show how to keep a balance between identity and attribute performance in the face swapping task.

\noindent\textbf{Efficient Id Embedding} The architecture of SimSwap uses the ID Injection Module to do the identity embedding so that we can separate the identity information from the weights of Decoder and generalize our architecture to arbitrary identity. To verify the effectiveness of our architecture, we conduct the same quantitative experiment on FaceForensics++~\cite{DBLP:faceforensics} using the criterion proposed by ~\cite{DBLP:FaceShifter}. We randomly pick 10 frames from each face video in FaceForensics++. We conduct face swapping using SimSwap following the same source and target pairs in FaceForensics++. We use another face recognition network~\cite{DBLP:cosface,CosFace_github} to extract the identity vectors of the generated frames and original frames. For each generated frame, we search for the nearest face in the original frames and check whether that face is from the correct source video. The accuracy rate is named as \textit{ID retrieval} and serves as a representation of the method's identity performance. We also use the pose estimator~\cite{DBLP:poseEstimation} to estimate the postures in the generated frames and the original frames, and we calculate their averaged L2 distance. We neglect the facial expression part because we are not able to find a valid reproduction. 

For further comparison, we train another 2 networks, called \emph{SimSwap-oFM} which uses the original Feature Matching formulation and \emph{SimSwap-nFM} which uses no Feature Matching term. We conduct the same quantitative experiments on these 2 networks. We also test on the frames generated by DeepFakes. The comparison results are shown in Table \ref{tab:faceforensics}.

As we can see, SimSwap-oFM has the lowset \textit{ID retrieval} since it aligns the results at shallow levels. Meanwhile, by removing the constrains at all levels, SimSwap-nFM has a very close identity performance as Faceshifter. Our SimSwap is a little behind in identity but achieves a relatively good posture performance. Combining with the results in Figure \ref{fig:DeepFakes} and \ref{fig:Faceshifter}, our SimSwap presents a slightly weaker identity performance than FaceShifter but possess a better attribute preservation ability.


\noindent\textbf{Keeping a Balance between Identity and Attribute} Since our IIM directly works on the whole feature $Fea_{T}$ extracted from target image, the embedding of identity will inevitably influence the performance of attribute preservation. So we need to find a balance between identity modification and attribute preservation.

 In our framework, there are two ways to adjust the balance between identity and attribute. The first is to explicitly set a heavier weight for $\lambda_{Id}$ to encourage a stronger modification skill. The second is to select more or less features in the Feature Matching term. The combination of these two approaches can result in a wide range of results.

Apart from SimSwap-oFM and SimSwap-nFM, we train another 4 networks, called \emph{SimSwap-$\overline{wFM}$}, \emph{SimSwap-oFM-FM-}, \emph{SimSwap-oFM-id+} and \emph{SimSwap-wFM-id+}. For $\overline{wFM}$, we keep the first few layers in the original Feature Matching term while removing the last few. For oFM-FM-, we utilize the original Feature Matching formulation and decrease $\lambda_{oFM}$ to 5. For oFM-id+, we utilize the original Feature Matching formulation and increase $\lambda_{Id}$ to 20. For wFM-id+, we keep the same Weak Feature Matching Loss as SimSwap and increase $\lambda_{Id}$ to 20. We train all the above networks following the same training strategy as SimSwap. We test the ID retrieval of all the networks.

\begin{table}
  \caption{Qualitative Experiments on FaceForensics++}
  \label{tab:faceforensics}

  \begin{tabular}{@{\hspace{.2cm}}c@{\hspace{1cm}}c@{\hspace{1cm}}c@{\hspace{.2cm}}}
    \toprule
    Method & ID retrieval & Posture \\ \hline
    DeepFakes~\cite{DeepFakes} & 77.65\% & 4.59\\
    FaceShifter~\cite{DBLP:FaceShifter} & \textbf{97.38\%} & 2.96\\ \hline
     SimSwap-oFM & 73.64\% & \textbf{1.22}\\
    SimSwap & 92.83\% & 1.53\\
    SimSwap-nFM & 96.57\% & 2.47\\
    \bottomrule

\end{tabular}
\end{table}

\begin{figure}
\centering
\includegraphics[width=8.5cm]{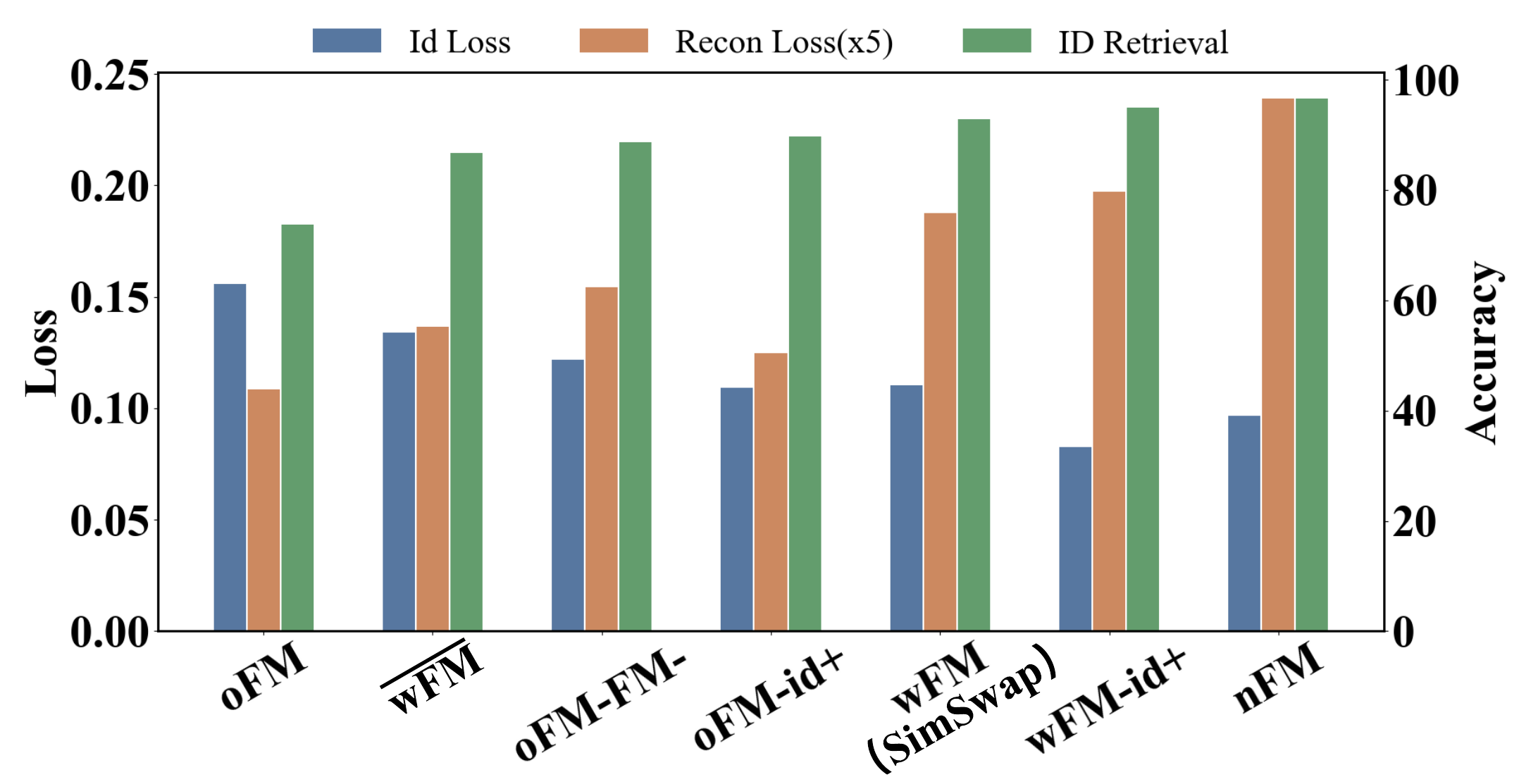}
\caption{Comparison between different Feature Matching term and Id weights.}
\label{fig:Ablation}
\end{figure}

We conduct additional quantitative experiments on CelebAMask-HQ~\cite{DBLP:MaskGAN,DBLP:CelebA}. First, we randomly pick 1,000 source and target pairs with different identities and generate face swapping results. We use the average Identity Loss between the result and the source to measure the identity modification skill. Then we randomly pick 1,000 images and use each image as both source and target to do self-swapping. We use the average Reconstruction Loss to measure how much attribute information from target has been lost during the face swapping process. The comparison results are shown in Figure \ref{fig:Ablation}.

\begin{figure*}[!ht] 
\centering
\includegraphics[width=1\textwidth]{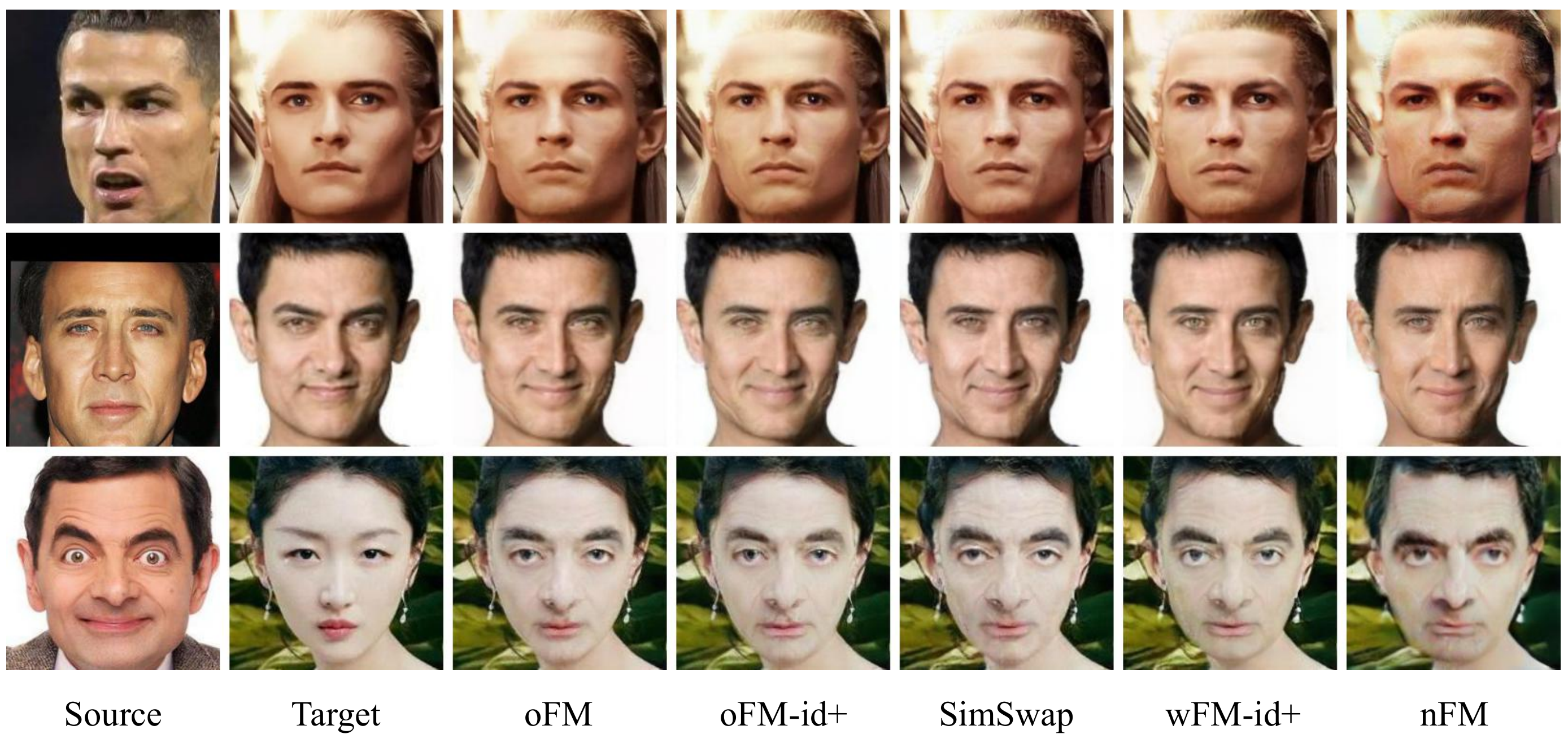}
\caption{Face swapping results generated by different networks. The results with original Feature Matching formulation(col 3\&4) exhibit a shortage of identity performance. The results with no Feature Matching term(col 7) have the best identity performance but suffer from attribute mismatch. The results with Weak Feature Matching Loss(col 5\&6) show a good balance between identity modification and attribute preservation.}
\label{fig:FaceShape}
\end{figure*}

As we can see, the oFM, $\overline{wFM}$, oFM-FM- and oFM-id+ all have lower ID retrieval than SimSwap. This indicates that keeping the last few layers in the Feature Matching term has less effect on the identity performance. Meanwhile, although nFM has the highest ID retrieval, its Reconstruction Loss is the highest among all, which shows that a strong modification ability will cause difficulty in attribute preservation. SimSwap is achieving a relatively high ID retrieval while keeping a medium Reconstruction Loss, which keeps a good balance between the identity and attribute performance.

\begin{figure}
\centering
\includegraphics[width=8.5cm]{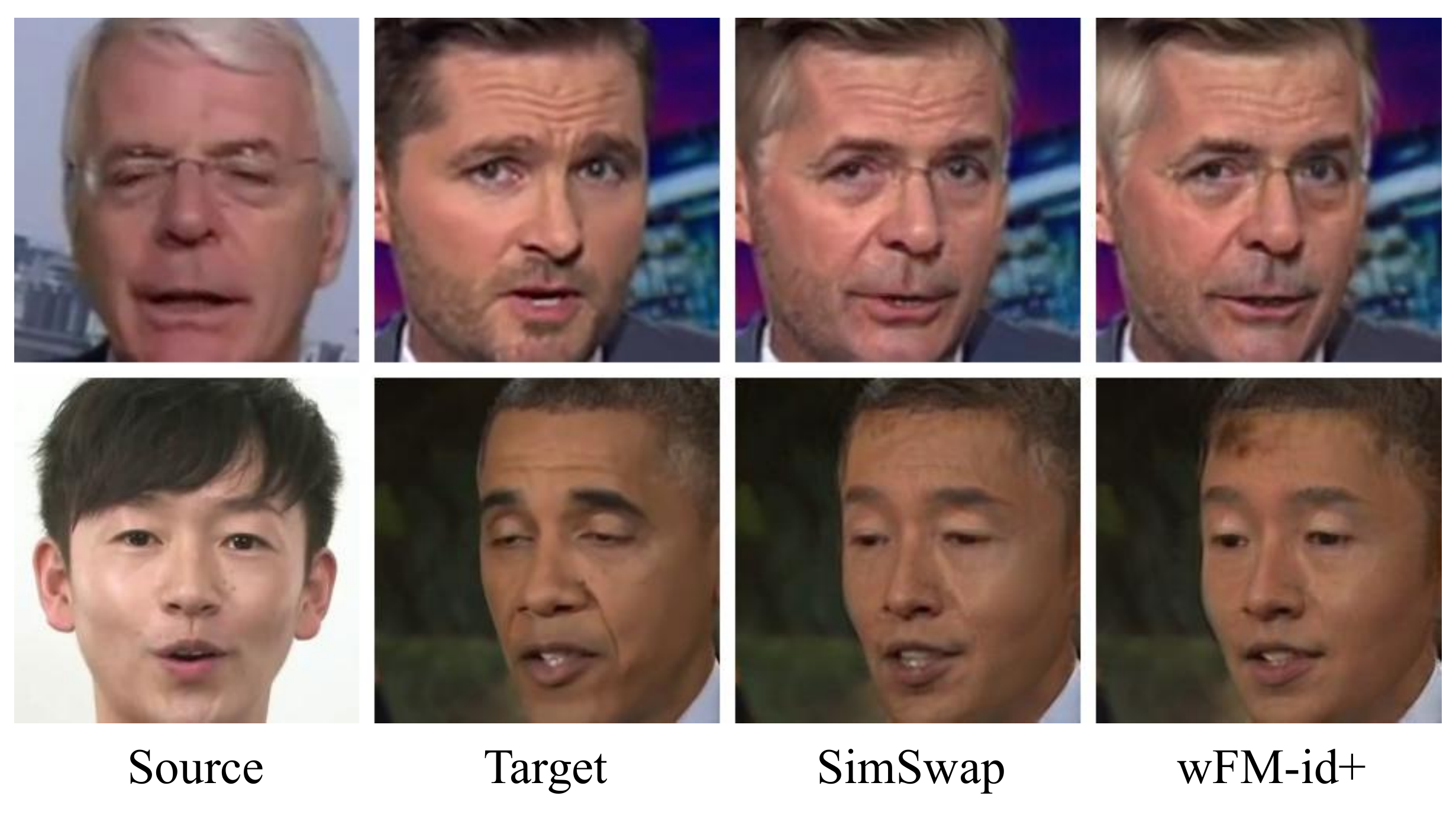}
\caption{Comparison between SimSwap and SimSwap-wFM-id+. SimSwap-wFM-id+ is more likely to introduce in hair from the source face.}
\label{fig:SS}
\end{figure}

To further validate the effectiveness of our Weak Feature Matching Loss, results generated by different networks are shown in Figure \ref{fig:FaceShape}. We notice that results(col 3\&4, col 5\&6) with the same Feature Matching term present relatively small difference, which indicates that the increase of $\lambda_{Id}$ has limited influence on visual looking. As we compare the results between SimSwap (col 5) and SimSwap-oFM (col 3), SimSwap is presenting a better identity performance without losing much attributes. The results of SimSwap-nFM (col 7) have the best identity performance, and the shapes of the result faces have been modified towards the shape of the source face. However, SimSwap-nFM is apparently losing attributes since gaze directions tend to deviate from those in the target faces.

As for SimSwap and wFM-id+, at most times they produce very similar visual outputs. However, when comparing the values of wFM-id+ and nFM in Figure \ref{fig:Ablation} and the results in Figure \ref{fig:FaceShape}, we notice that although wFM-id+ has a smaller Identity Loss, the ID retrieval and visual looking show that the nFM achieves better identity performance indeed. This indicates that wFM-id+ has been overfitted for the Identity Loss. Besides, wFM-id+ is more likely to introduce in hair from the source face (as shown in Figure \ref{fig:SS}). This is unwanted since we are only replacing the faces. So we are choosing SimSwap for more stable results.

\section{Conclusion}

We propose \emph{SimSwap}, an efficient framework aiming for generalized and high fidelity face swapping. Our ID Injection Module which transfers the identity information at feature level and extends the identity-specific face swapping to arbitrary face swapping. The Weak Feature Matching Loss helps our framework to possess a good attribute preservation ability. Extensive results have shown that we are able to generate visually appealing results and our method can preserve attributes better than previous methods.

\begin{acks}
This work was supported by National Science Foundation of China (61976137, U1611461, U19B2035) and STCSM(18DZ1112300).
\end{acks}

\bibliographystyle{ACM-Reference-Format}
\bibliography{sample-base}

\appendix

\section{Additional Comparison with DeepFakes}

\begin{figure*}[!ht]
  \centering
  \includegraphics[width=1\linewidth]{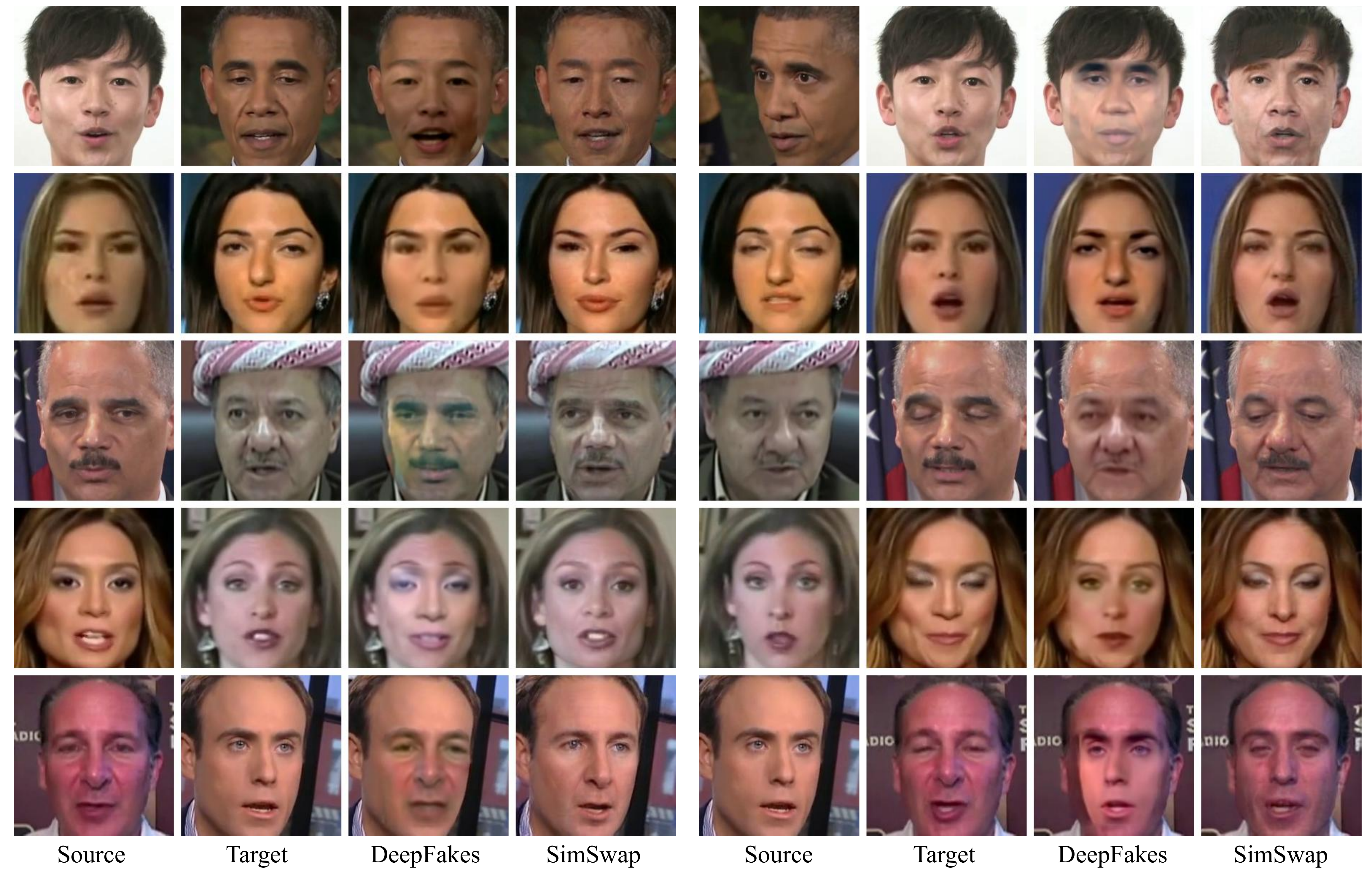}
  \caption{More comparison results on FaceForensics++~\cite{DBLP:faceforensics}. Even though we have not trained on source and target faces, we manage to produce better results than DeepFakes~\cite{DeepFakes}.}
  \label{fig:deepfakes_more}
\end{figure*}

We show more comparison with DeepFakes~\cite{DeepFakes} on FaceForensics++~\cite{DBLP:faceforensics} in Figure \ref{fig:deepfakes_more}.

\section{Video Results}

We provide 5 face swapping videos generated by SimSwap. The name format is s\_X\_t\_Y.avi, where X stands for the source identity and Y stands for the target identity. We put the source image on the upper left corner and the target image on the bottoem left corner.  Please check our videos for details.
The video is available on our github project.
\section{Additional Results for SimSwap}

\begin{figure*}[!ht]
  \centering
  \includegraphics[width=1\linewidth]{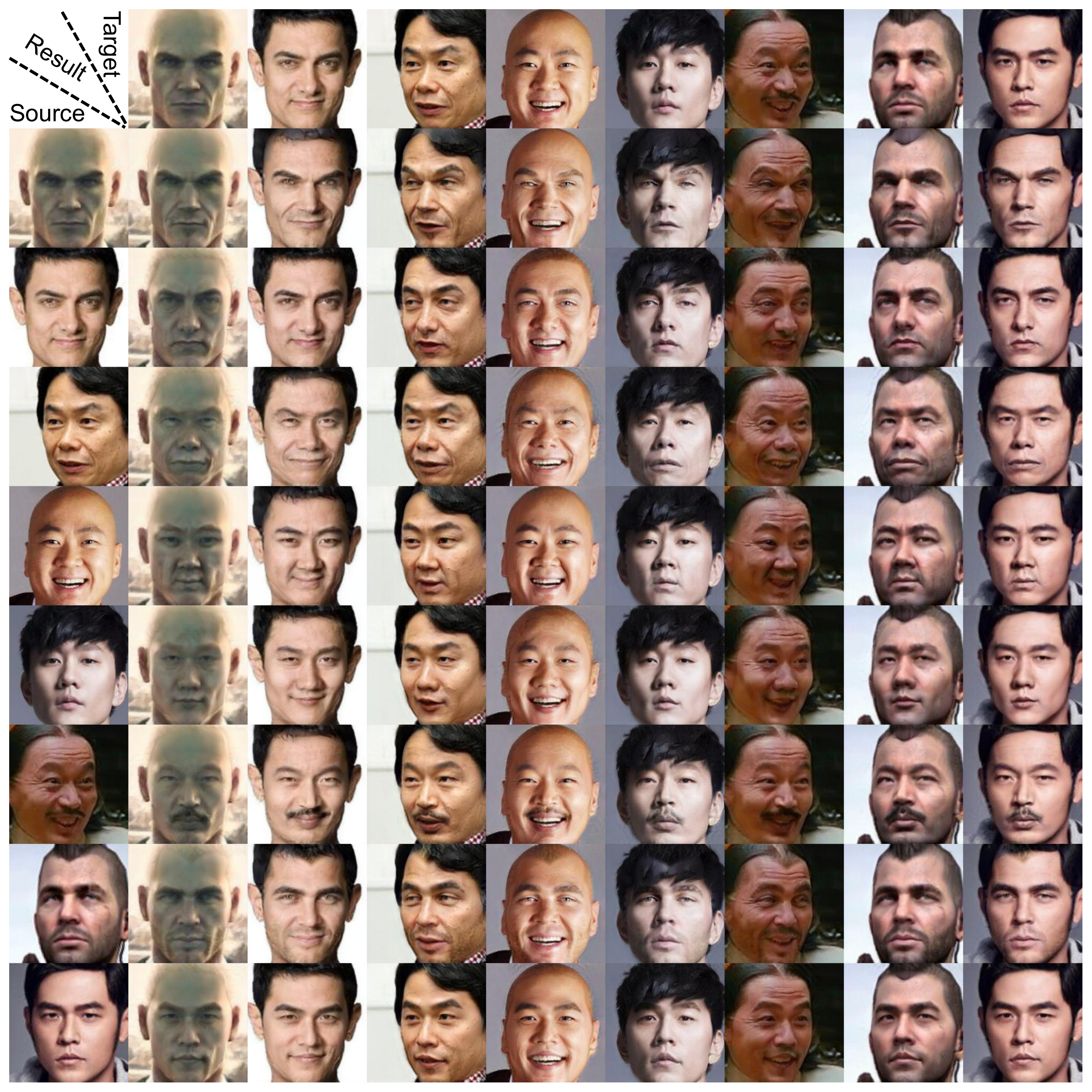}
  \caption{Male Face Matrix generated by SimSwap. The identities in row 1 and row 7 are fictional characters. Our method manages to generate high fidelity face swapping result.}
  \label{fig:matrix_sup_1}
\end{figure*}

\begin{figure*}[!ht]
  \centering
  \includegraphics[width=1\linewidth]{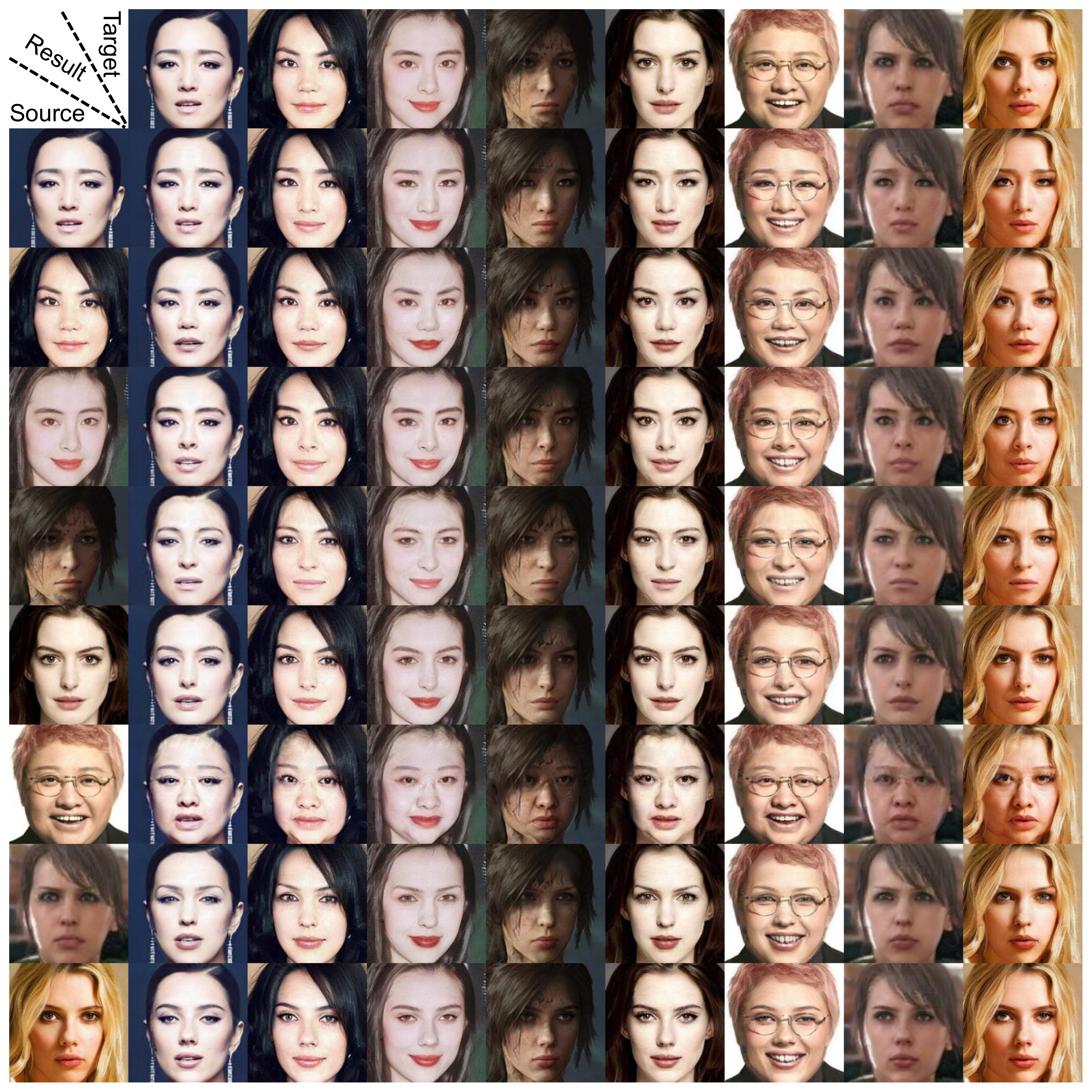}
  \caption{Female Face Matrix generated by SimSwap. The identities in row 4 and row 7 are fictional characters. Our method manages to generate high fidelity face swapping result.}
  \label{fig:matrix_sup_2}
\end{figure*}

\begin{figure*}[!ht]
  \centering
  \includegraphics[width=1\linewidth]{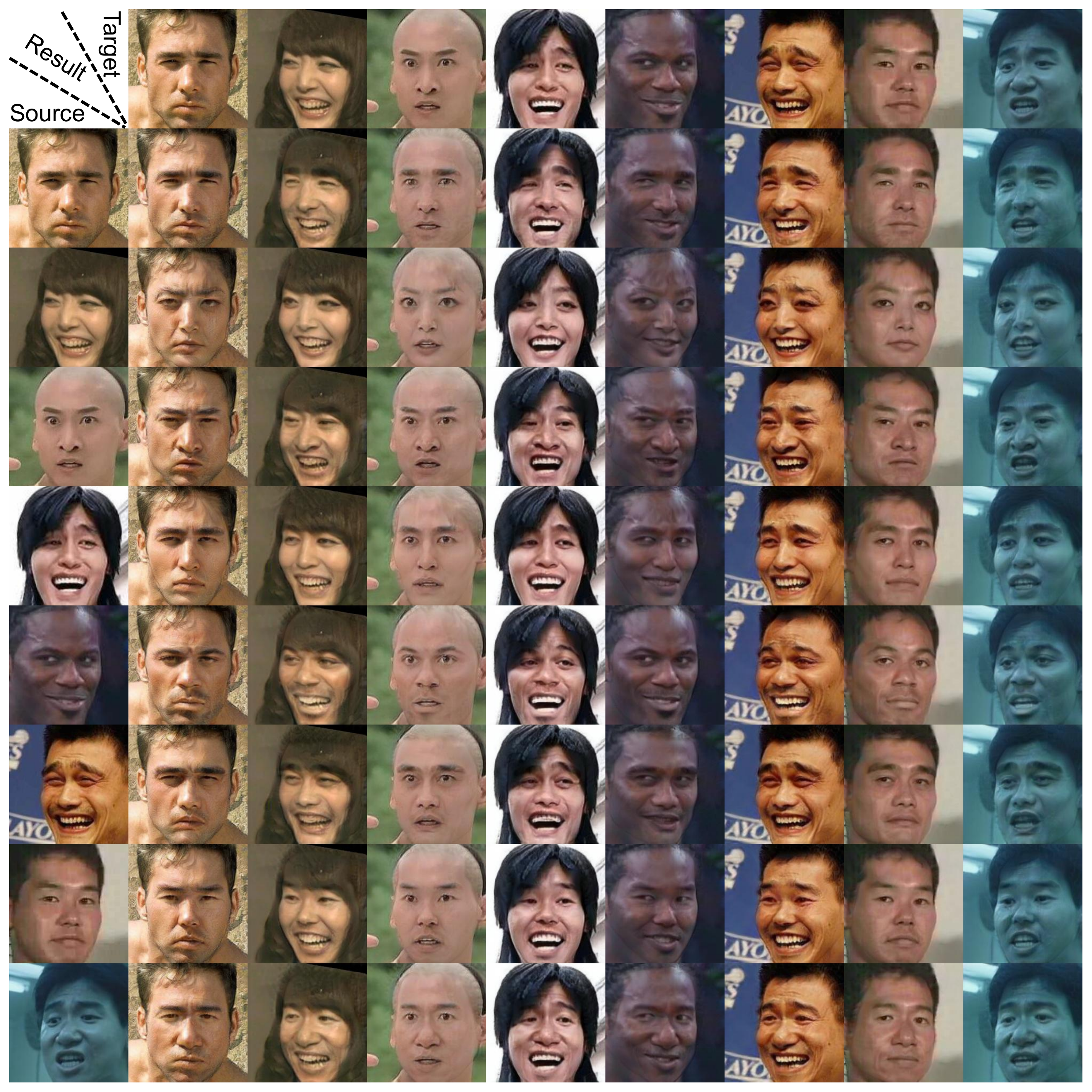}
  \caption{Expression Face Matrix generated by SimSwap. Some identities in this matrix have extraggerated expressions. Our method is still to generate decent face swapping result.}
  \label{fig:matrix_sup_3}
\end{figure*}

We show more face matrix generated by SimSwap in Figure \ref{fig:matrix_sup_1},\ref{fig:matrix_sup_2},\ref{fig:matrix_sup_3}. We even include some fictional characters.

\end{document}